\documentclass[journal,twoside,web]{ieeecolor}
\usepackage{generic}
\usepackage{cite}
\usepackage{amsmath,amssymb,amsfonts}
\usepackage{graphicx}
\usepackage{textcomp}
\usepackage{makecell}
\usepackage{multirow}
\usepackage{pifont}
\usepackage{hyperref}
\usepackage{soul}
\soulregister\cite7 
\usepackage{xcolor}
\setul{0.5ex}{2pt}
\setstcolor{red}
\usepackage{soul}
\soulregister\ref7
\soulregister\pageref7
\usepackage[normalem]{ulem}

\def\BibTeX{{\rm B\kern-.05em{\sc i\kern-.025em b}\kern-.08em
    T\kern-.1667em\lower.7ex\hbox{E}\kern-.125emX}}
\begin{document}
\bstctlcite{bstctl:nodash}
\title{Simulation-Driven Imitation Learning for Biosignals-Free Shared-Autonomy Prosthetic Grasping}
\author{
Kaijie Shi, Wanglong Lu, Huiling Chen, \IEEEmembership{Associate Member, IEEE}, Vinicius Prado da Fonseca, \IEEEmembership{Member, IEEE}, Ting Zou, \IEEEmembership{Senior Member, IEEE}, Hanli Zhao\MakeUppercase{*}, and Xianta Jiang\MakeUppercase{*}, \IEEEmembership{Senior Member, IEEE}
\thanks{\textcolor{red}{This work has been submitted to the IEEE for possible publication. Copyright may be transferred without notice, after which this version may no longer be accessible.}}
\thanks{Submitted on \today. This work was supported in part by the Government of Canada's New Frontiers in Research Fund (NFRF, Grant No NFRFE-2022-00407) and Natural Sciences and Engineering Research Council of Canada's Research Tools and Instruments (NSERC RTI, Grant No  RTI-2022-00688).}
\thanks{This work involved human subjects or animals in its research. Approval of all ethical and experimental procedures and protocols was granted by the Memorial University Interdisciplinary Committee on Ethics in Human Research (20210316-SC).}
\thanks{
Kaijie Shi and Wanglong Lu are with Department of Computer Science, Memorial University of Newfoundland, St. John’s, NL A1B 3X5, Canada, and also with College of Computer Science and Artificial Intelligence, Wenzhou University, Wenzhou, 325000, China. (email: kaijies@mun.ca, wanglongl@mun.ca).
}
\thanks{
Huiling Chen and Hanli Zhao are with College of Computer Science and Artificial Intelligence, Wenzhou University, Wenzhou, 325000, China. (email: chenhuiling\_jsj@wzu.edu.cn, hanlizhao@wzu.edu.cn).
}
\thanks{
Vinicius Prado da Fonseca and Xianta Jiang are with Department of Computer Science, Memorial University of Newfoundland, St. John’s, NL A1B 3X5, Canada. (email: vpradodafons@mun.ca, xiantaj@mun.ca).
}
\thanks{Ting Zou is with Department of Mechanical and Mechatronics Engineering, Memorial University of Newfoundland, St. John’s, NL A1B 3X5, Canada. (email: tzou@mun.ca). }
\thanks{*Corresponding authors.}
}

\maketitle

\begin{abstract}
Biosignals-free shared-autonomy control of upper-limb prosthetic hands aims to enable natural and low-effort manipulation without relying on EMG or other physiological signals.
Recent imitation-learning-based approaches have shown promising results, but their scalability is limited by the cost and variability of collecting large amounts of real-world human demonstration data.
In this work, we present a scalable simulation framework that automatically generates diverse reach-to-grasp demonstrations from a wrist-mounted virtual camera.
The framework combines physically feasible grasp synthesis, natural reaching trajectories retargeting, and reach--grasp--lift execution in procedurally generated  indoor environments. 
It records wrist-view observations, proprioception, and actions to build a large-scale demonstration dataset for imitation learning.
Through extensive simulation benchmarks, we evaluate object and scene generalization and compare several representative state-of-the-art imitation learning methods. Results show that the simulated demonstrations are sufficiently rich and consistent for effective policy learning.
In three realistic settings, the learned sim-to-real policy achieves over 90\% grasp success, surpasses baseline methods, and exhibits stronger generalization, highlighting the promise of simulation-driven training for biosignals-free shared-autonomy prosthetic grasping. 
The demonstrations are available at \href{https://sites.google.com/view/sim-prosthetic-grasp/home}{https://sites.google.com/view/sim-prosthetic-grasp/home}.


\end{abstract}


\section{Introduction}

The loss of a hand profoundly limits an individual’s ability to interact with objects in daily life. Prosthetic hands provide a non-invasive means to restore lost functionality, offering users the ability to perform basic grasping tasks. However, how to control these prostheses in an efficient, intuitive, and reliable manner remains a fundamental challenge \cite{dhawan2019proprioceptive, farina2014extraction, cipriani2011online}.

Current mainstream prosthetic hand control methods are primarily based on surface electromyography (sEMG) signals \cite{roche2014prosthetic, nasr2021musclenet, boostani2003evaluation, dalley2011method}. In these approaches, users generate distinct muscle signal patterns to trigger specific grasp types. While widely studied, sEMG-based methods place a heavy burden on the user \cite{farrell2007optimal, wang2021effect, fang2022simultaneous}, as they require active participation at every step of the grasping process. To alleviate this limitation, semi-autonomous approaches have been proposed, in which computer vision is used to identify the desired grasp type, and sEMG signals are then employed to initiate each motion \cite{dovsen2010cognitive, he2020vision}. Although such methods reduce some of the decision-making burden, they still require explicit per-grasp triggering, which can be physically and psychologically demanding.

To overcome these limitations, autonomous grasping has recently attracted significant attention \cite{shi2025towards, alessi2025hannesimitation}. The system infers user intent directly from natural limb positioning and a brief dwell near the target object, while autonomously executing low-level grasp and manipulation actions mimicking human dexterity. The key idea is to shift the low-level decision-making process from the user to the prosthesis itself, thereby simplifying control. For instance, Shi et al. \cite{shi2025towards} proposed a biosignals-free autonomous control method that eliminates sEMG dependency altogether. 
This approach relies solely on a wrist-mounted camera to perceive object properties and autonomously execute grasp-and-release actions. In this framework, users are only responsible for moving the prosthesis to the desired location, while the system independently manages grasping and releasing. Beyond reducing user effort, such biosignals-free methods also offer strong cross-user generalization, since they do not require subject-specific data collection and model training \cite{shi2025towards}.

Nevertheless, autonomous grasping introduces new challenges, particularly in visual generalization. Vision-based imitation learning models often degrade when facing distribution shifts such as novel objects, cluttered backgrounds, or changes in illumination \cite{gupta2018robot, tobin2017domain, hsu2022vision}. A common way to improve robustness is to learn policies from a large-scale and diverse dataset \cite{kim24openvla}. However, it is fundamentally bottlenecked by the cost of collecting sufficiently large and diverse demonstration datasets. This bottleneck becomes especially severe for real-world upper-limb prosthetic grasping: demonstrations must be recorded on physical hardware under human-in-the-loop operation, often requiring careful setup and calibration, repeated trials to cover object and pose variability, and strict safety considerations when interacting with users and everyday environments. As a result, data collection is labor-intensive, slow to collect, and difficult to scale; operator fatigue and day-to-day changes in execution can further introduce inconsistency and noise. Moreover, practical constraints such as hardware availability, wear-and-tear, and limited experimental time make it difficult to capture the long-tail cases that are critical for robust grasping in the wild.

To address this data collection bottleneck, we propose a simulation framework for biosignals-free prosthetic hand grasping. The framework enables automatic generation of diverse and scalable training data by simulating key factors in grasping, including wrist trajectories, grasp configurations, objects, and background environments. By systematically varying these components, the system can, in principle, generate an infinite number of prosthetic hand grasping scenarios. Compared with prior work on virtual simulation (e.g., Xie et al. \cite{xie2021virtual}, 2021), our framework is specifically designed to accelerate data generation for autonomous prosthesis grasping neural network training, thereby reducing reliance on labor-intensive real-world demonstration collection.


To the best of our knowledge, this is the first simulation framework served for biosignals-free shared-autonomy prosthetic grasping research area. 
This study presents the following key contributions:
\begin{enumerate}


\item \textbf{Demonstration Synthesis.}
We build a fully automated data collection pipeline in simulation to generate a scalable set of reach-to-grasp trials, covering diverse objects, wrist trajectories, and procedurally generated indoor scenes, with no manual intervention required.

\item \textbf{Imitation-learning Benchmarking. }
We establish a standardized and reproducible simulation benchmark for autonomous prosthetic grasping, filling a gap in evaluation protocols and enabling systematic comparison of representative SOTA imitation-learning methods under multiple generalization settings.

\item \textbf{Realistic-setting Validation.}
We establish the practical viability of simulation-trained policies for autonomous prosthetic grasping by transferring them to a real prosthetic hand and validating their performance across 1,800 real-world trials involving 12 participants.

\end{enumerate}

\begin{figure*}[htbp!]
  \centering  \includegraphics[page=1,width=\linewidth]{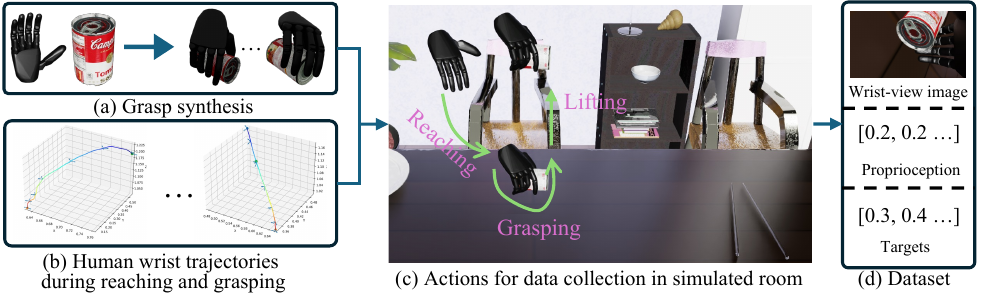}
  \caption{Overview of the proposed simulated data collection framework. Given an object and a hand model, we first synthesize feasible grasps (a). A human wrist trajectory for reaching and grasping (b) is then sampled and retargeted to the synthesize grasp before being executed in the simulation environment (c). During this process, wrist-view images, proprioceptive measurements and action targets are recorded to build the dataset (d).}
  \label{fig:teaser}
\end{figure*}

\section{Related work}

\subsection{Control paradigms for prosthetic grasping}
Mainstream prosthetic-hand control has long relied on surface electromyography (sEMG), where users produce distinct muscle-activation patterns to trigger predefined grasp types or action sequences \cite{roche2014prosthetic,nasr2021musclenet,boostani2003evaluation,geethanjali2014low, patel2018classification, hahne2015concurrent}. Despite extensive study and deployment, sEMG-based control can impose substantial user burden because effective operation often requires active participation and explicit triggering throughout the grasping process \cite{jiang2012myoelectric,wang2021effect,fang2022simultaneous, jiang2013intuitive}.

To reduce decision-making effort, semi-autonomous approaches use computer vision to infer candidate grasp types or target properties, while sEMG is used to confirm or initiate motions \cite{dovsen2010cognitive,he2020vision, xu2025powered}. Although such hybrids partially offload perception and selection, they frequently retain per-grasp confirmation/triggering steps, which can remain physically and psychologically demanding during repeated daily interactions.

Recently, \emph{autonomous grasping} has emerged as a promising direction to further reduce user burden by shifting ``what to grasp, how to grasp, and when to grasp'' decisions from the user to the prosthesis controller \cite{shi2025towards,alessi2025hannesimitation}. In particular, \emph{biosignals-free} methods eliminate sEMG dependency altogether, avoiding subject-specific signal collection and calibration and potentially improving cross-user generalization. Representative work leverages a wrist-mounted camera to perceive objects and autonomously execute grasp-and-release behaviors, while the user primarily positions the prosthesis \cite{shi2025towards}. Imitation-learning-based approaches for prosthetic grasping have also been explored \cite{alessi2025hannesimitation}. 

Beyond vision-driven autonomy, the literature includes non-vision (or weak-vision) semi-autonomous paradigms that exploit residual-limb kinematics and postural synergies, inertial sensing, proximity sensing and sensor fusion, and adaptive mode-switching to reduce control dimensionality or simplify interaction \cite{montagnani2015exploiting,kuhn2024synergy,merad2020assessment,bennett2017imu,pilarski2012dynamic,mastinu2024explorations,heo2023proximity}. While these approaches offer valuable alternatives, they primarily emphasize interface design, intent inference, or sensor configurations, and are not directly targeted at scalable learning and benchmarking for wrist-vision autonomous grasping, which is the focus of this work.

\subsection{Simulation for grasping}
Collecting large-scale grasping data on real robotic hands is expensive, time-consuming, and potentially damaging to hardware. Simulation provides a scalable alternative by enabling automatic generation of grasp scenes, object interactions, and success labels with precise annotations \cite{fang2020graspnet, chen2025bodex, wang2022dexgraspnet, liu2024structured, yu2025trustworthy, xu2025anthropomorphic,wang2025toward}. Prior work has used physics simulators to model object poses, contact dynamics, and grasp outcomes, while photorealistic rendering and procedural scene generation have been employed to synthesize diverse visual observations. Domain randomization further improves diversity by varying textures, lighting, camera parameters, and physical properties, making simulation a practical source of training data for grasp learning \cite{zhang2024dexgraspnet, tobin2017domain, tobin2018domain, huber2024domain}.

Imitation learning has been widely adopted to learn grasping policies from expert demonstrations, which may come from human teleoperation \cite{zhao2023learning, kim24openvla}, motion planners \cite{mandlekar2023human, choudhury2018data, zare2024survey}, or simulated experts \cite{chi2023diffusion}. Behavior cloning \cite{torabi2018behavioral} is a common approach that directly maps visual observations to grasp poses, end-effector actions, or gripper commands. Although effective and easy to optimize, such methods often suffer from compounding errors caused by distribution shift between training and deployment \cite{ross2011reduction}. To improve robustness, interactive imitation learning and data aggregation methods have been explored to provide corrective supervision on states visited by the learner \cite{kelly2019hg}. Recent work further combines imitation learning with deep visual representations and sequence models for closed-loop grasping and long-horizon manipulation tasks \cite{levine2016end, kim24openvla, zhao2023learning}.

A major challenge in simulation-trained grasping is the sim-to-real gap, which arises from discrepancies in visual appearance, depth noise, contact dynamics, friction, actuation, and calibration errors. Since grasp success is highly sensitive to both perception and contact modeling, these discrepancies can substantially degrade real-world performance. Existing approaches \cite{james2019sim, tobin2017domain, patel2025real} address this issue through visual domain randomization, domain adaptation, dynamics randomization, and system identification. Some methods further incorporate real-world fine-tuning, residual policy learning, or online adaptation after transfer \cite{ju2022transferring, ankile2025residual}. These advances have significantly improved the feasibility of deploying simulation-trained grasping policies on real robotic systems.

\section{Simulation framework}



We aim to automatically generate scalable reach-to-grasp demonstration data for training an imitation-learning policy for biosignal-free prosthetic grasping. Each demonstration contains wrist-view RGB observations and proprioceptive states, together with joint-level action supervision for learning dexterous finger control.

Figure~\ref{fig:teaser} illustrates our framework for generating scalable simulated demonstrations. Given a set of objects, a prosthetic hand model, and indoor scenes in simulation (Section~\ref{sec:objects and hand}), we first synthesize diverse grasps for each object (Section~\ref{sec:Grasp synthesis}). We then sample a natural wrist-reaching motion and retarget its final wrist pose to match the target grasp (Section~\ref{sec: trajectory}). Finally, we execute the reaching, grasping, and lifting sequence in simulation (Section~\ref{sec: Dataset construction}), while recording wrist-camera images, proprioception, and joint actions to construct the dataset for training autonomous grasping models.

\begin{figure}[tbp!]
  \centering  \includegraphics[page=1,width=\linewidth]{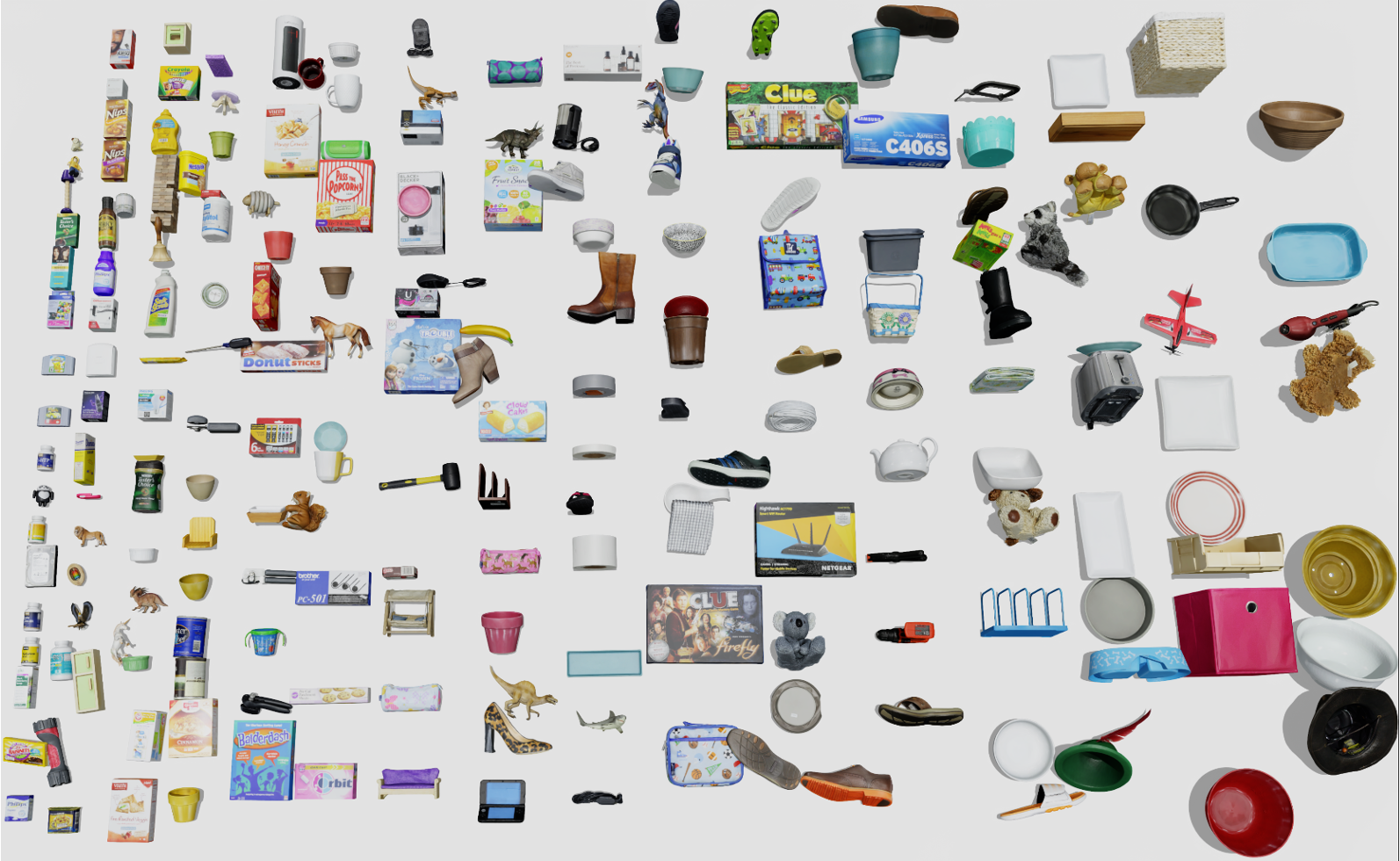}
  \caption{200 objects used in this paper.}
  \label{fig:objects}
\end{figure}

\subsection{Objects and Prosthetic hand} \label{sec:objects and hand}
\textbf{Objects.}
As shown in Figure~\ref{fig:objects}, 200 objects are used in this paper, sampled from the Google Scanned Objects (GSO) dataset~\cite{downs2022google}. GSO is an open-source collection of over one thousand high-quality 3D-scanned household items released under a Creative Commons license, providing textured, realistic meshes that are preprocessed for common physics simulators (e.g., Ignition Gazebo, Bullet), which makes it well-suited for interactive simulation, synthetic data generation, and robotic perception and manipulation tasks. Furthermore, we use the Universal Scene Description (USD) assets processed following Multigrippergrasp \cite{casas2024multigrippergrasp}, enabling direct use of these objects in NVIDIA Isaac Sim, which is the main simulation utility used in this work.

\textbf{Prosthetic Hand.}
In this study, the prosthetic hand (Ability hand, PSYONIC \cite{psyonic_touch_sensing_bionic_hand_2024}) features six degrees of freedom (DoFs): one for each finger, with the thumb having two DoFs. We use the 3D hand model provided by PSYONIC company in the simulation.
While this study uses the Ability Hand, the approach is largely hardware-agnostic and can be transferred to other multi-DOF prosthetic hands with minimal changes, primarily in action-space definition, kinematic retargeting, and sensor calibration, provided that a suitable 3D model of the target hand is available for simulation.

\textbf{System Identification.}
We calibrate joint dynamics in simulation via system identification so that the same joint command produces similar motion in simulation and on the real prosthetic hand, reducing the action gap for sim-to-real transfer.
Specifically, we first record a sequence of target action - proprioception pairs on the real prosthetic hand, and then use an evolutionary algorithm in simulation to search for the stiffness and damping settings of each joint that best reproduce this behavior.

\textbf{Simulated Wrist Camera.} 
In simulation, an RGB camera is mounted on the prosthetic wrist and oriented toward the workspace to capture hand–object interactions. The camera extrinsics are defined relative to the hand base frame, and the intrinsics (resolution, focal length/FOV) are set to match the real wrist camera (RealSense D405); images are rendered at $424 \times 240 $.

\begin{figure}[tbp!]
  \centering
  \includegraphics[page=1,width=\linewidth]{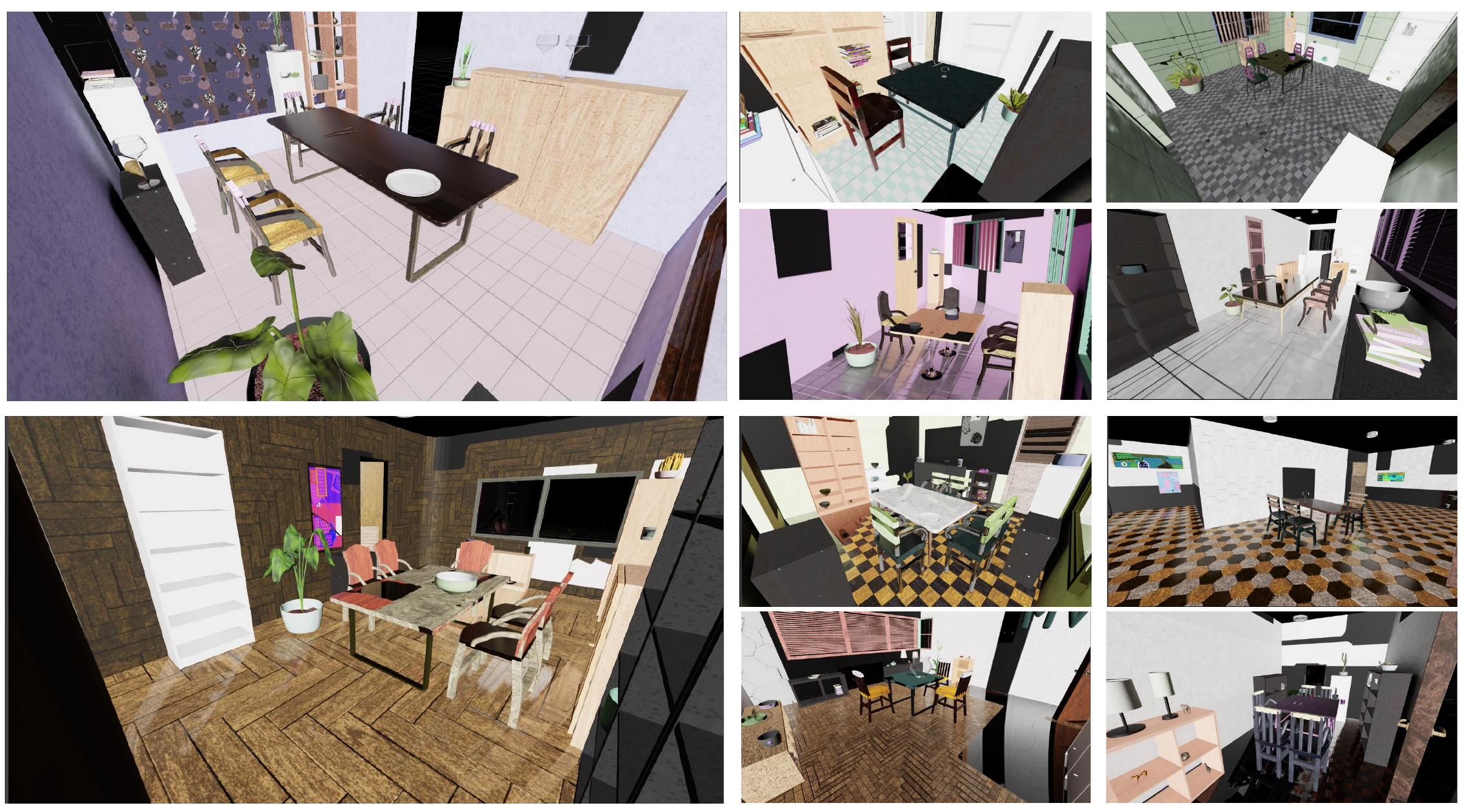}
  \caption{
  Wide-angle corner views of ten indoor rooms were generated with Infinigen Indoors \cite{raistrick2024infinigen} and used in our experiments.
  }
  \label{fig:scene}
\end{figure}

\textbf{Indoor Scenes.} We use Infinigen Indoors \cite{raistrick2024infinigen} to generate 10 indoor rooms for our experiments (Figure~\ref{fig:scene}). Infinigen Indoors procedurally synthesizes photorealistic indoor scenes by sampling from a large library of procedural assets (e.g., furniture, architectural elements, appliances, and everyday objects) and arranging them via a constraint-based layout system. A domain-specific language and a dedicated solver are employed to enforce both semantic coherence and physical plausibility in the resulting environments \cite{raistrick2024infinigen}.

\subsection{Grasp synthesis} \label{sec:Grasp synthesis}

We do not use hand-crafted grasp labels for each object (e.g., assigning a spherical grasp to a ball). Instead, we use synthetic grasps in place of human demonstrations, enabling faster data collection and greater grasp diversity. Following BoDex \cite{chen2024bodex}, we synthesize dexterous grasps via optimization in simulation for the prosthetic hand.

Given an object $O=(M_o, T_o)$, where $M_o$ is the object mesh and $T_{o}\in SE(3)$ denotes its pose, the prosthetic hand is represented as $H=(M_h, q_h)$, where $M_h$ is the hand mesh and $q_h$ is the initial finger joint configuration corresponding to an open-hand posture. The resulting grasp is parameterized as $g=(T_g, q_g)$, where $T_g$ is the wrist pose and $q_g$ is the optimized finger joint configuration corresponding to the grasping posture.

Compared to another grasp generator for prosthetic grasp, Bring Your Own Grasp Generator (BYOGG)~\cite{stracquadanio2025bring}, which generates parallel-jaw gripper grasps (e.g., via Contact-GraspNet \cite{sundermeyer2021contact}) and then maps them to a low-dimensional prosthetic pre-shape, BoDex~\cite{chen2024bodex} directly optimizes the dexterous grasp in the full configuration space (wrist pose plus all joint angles). Its bilevel formulation explicitly solves for frictional contact forces and validates grasps in simulation, producing contact-rich and physically feasible grasps that are better suited for large-scale synthetic supervision.

\subsection{Hand move trajectory} \label{sec: trajectory}

\textbf{Reaching trajectory patterns.} We maintain a database of reaching-phase wrist trajectories
$\{\tau_i\}_{i=1}^{N}$~\cite{wu2023learning}, where $N$ is the number of trajectories.
The wrist motion is obtained by manually annotating per-frame 2D hand keypoints (including the wrist) in multi-view RGB-D videos and then reconstructing the per-frame 3D wrist pose \cite{chao2021dexycb}.
Each trajectory $\tau_i$ is represented by a sequence of $m_i$ critical wrist poses
$\tau_i = \{T_{i,1}, \dots, T_{i,m_i}\}$, where $T_{i,j}\in SE(3)$ denotes the wrist rigid transform
(position and orientation) at the $j$-th keyframe.

\textbf{End pose retargeting.} We use the wrist pose of each synthetic grasp (Sec.~\ref{sec:Grasp synthesis}) as the desired terminal pose of the reaching trajectory, denoted by $T_G\in SE(3)$.
To retarget a template trajectory $\tau_i$ to end exactly at $T_G$, we compute the aligning transform
\begin{equation}
\Delta_i = T_G \, T_{i,m_i}^{-1},
\end{equation}
and apply it to every keyframe:
\begin{equation}
\hat{T}_{i,j} = \Delta_i \, T_{i,j}, \quad j=1,\dots,m_i.
\end{equation}
This construction guarantees $\hat{T}_{i,m_i} = T_G$, i.e., the final frame of the reaching phase
matches a physically feasible grasp wrist pose.

\begin{table}[htbp]
  \centering
  \caption{Reach-to-grasp execution.}
  \label{tab:wide}
  \begin{tabular}{|c|c|c|}
  \hline
    Time step index & Wrist action & Fingers action  \\ \hline 
    1-50 & Reaching & Keep Open \\ \hline
    51-65 & Stoping & Execute grasping \\ \hline
    66-115 & Lifting & Keep grasping \\ \hline
  \end{tabular}\label{tab:execution}
\end{table}

\subsection{Dataset construction} \label{sec: Dataset construction}

\textbf{Object pose randomization.}
At the beginning of each trial, we randomize the object pose by initializing it at a height of 1~meter with its planar position sampled as
$x,y \sim \mathcal{U}(-1,1)$ meter.
We then release the object and let it fall under gravity.
After it comes to rest, we read out its final pose $T_o^{\text{rest}}$ and transform the corresponding grasp accordingly, so that the grasp pose remains consistent with the settled object state.

\textbf{Reach-to-grasp execution.}
We execute each grasp episode for $115$ time steps (1 second has 30 time steps) and divide the execution into three phases (Table \ref{tab:execution}).
\begin{enumerate}
\item \emph{Reaching} ($t=1$--$50$): the wrist follows the reaching trajectory toward the target pre-grasp pose, while the fingers remain fully open to avoid premature contact.
\item \emph{Stopping \& Grasping} ($t=51$--$65$): the wrist motion is decelerated and brought to a stop at the terminal grasp wrist pose; meanwhile, the fingers execute the grasp by closing to establish stable contacts.
\item \emph{Lifting} ($t=66$--$115$): the wrist lifts the object upward, and the fingers maintain the grasp configuration to keep the object secured throughout the lift.
\end{enumerate}

\textbf{Trial collection.} 
With 200 objects and 10 rooms, we collected 2000 successful reach-to-grasp trials.
Each trial lasts for 115 time steps, and at each step we record (i) an RGB image from the wrist-mounted camera,
(ii) proprioception (per-joint angles), and (iii) the target joint position for each finger.
A trial is labeled as successful if the object remains stably grasped during lifting, quantified by a small change in the
object-to-hand relative position between lift onset and the end of the episode:
$\left\| (p_{\mathrm{obj}}-p_{\mathrm{hand}})^{\mathrm{final}} - (p_{\mathrm{obj}}-p_{\mathrm{hand}})^{\mathrm{lift}} \right\|_2 < 0.05\,\mathrm{m}$.
We split the 2000 successful trials into training and testing sets for generalization experiments; the split protocol is described in Sec.~\ref{sec: simulation experiments}.

\textbf{Demonstrations quality control.} We ensure demonstration quality by using physically feasible synthesized grasps and retaining only trials that satisfy a stability-based success criterion during lifting.

\section{Simulation Experiments} \label{sec: simulation experiments}
\subsection{Experimental setup.}
All simulation experiments are conducted in Isaac Lab.
We consider 200 objects and 10 procedurally generated rooms in the simulation.
We first collect 2000 successful reach-to-grasp trials and construct training/testing splits for different generalization settings (Sec.\ref{sec: sim_generalization}).
At test time, each grasp trial is executed in 5 parallel environments, and we report the aggregated performance over these runs.
We focus on this scale to provide a controlled benchmark; scaling up is left for future work.

\subsection{Imitation learning benchmark algorithms}





\textbf{Action Chunking with Transformers (ACT)~\cite{zhao2023learning}.}
ACT employs a CVAE-Transformer to model the conditional distribution over image and proprioception, and predicts an action chunk (multiple future actions) at each step, optionally using temporal ensembling for smooth execution.
We adapt ACT to our prosthetic setting by conditioning it on the wrist-view RGB observation and proprioception, and mapping its predicted action chunks to multi-step joint command sequences for the prosthetic hand. 

\textbf{VTM-VAE~\cite{shi2025towards}.}
VTM-VAE is a variational model for prosthetic control and replaces the Transformer with a central-aware Mamba backbone compared with ACT to reduce computation for image understanding.
Since tactile sensing is not available in our setup, we remove the tactile stream and condition the model on wrist-view RGB and proprioception only, while mapping the outputs to prosthetic joint command sequences. 

\textbf{HannesImitation~\cite{alessi2025hannesimitation}.}
HannesImitation proposes \emph{HannesImitationPolicy}, a diffusion-policy-based imitation learning pipeline to control a prosthetic hand for grasping from eye-in-hand RGB and proprioceptive signals~\cite{alessi2025hannesimitation}.
To benchmark a diffusion-based prosthetic-specific approach under our setting, we adopt its diffusion-policy backbone and interface it with our observation space (wrist-view RGB and proprioception) and action space (prosthetic joint commands). 

\textbf{Implementation details.}
For those three SOTA imitation-learning methods, we change the algorithm's input and output, and keep the model's backbone same as original paper.

For ACT and VTM-VAE, the observation consists of (i) the current wrist-view RGB image and (ii) a proprioceptive window of joint positions over the past 30 steps (current + 29 history).
The policy outputs a horizon of 25 future joint command vectors.
The image size is $320 \times 240$, while the size of joint state and target pose/angle at a specific time step is 6.
We train neural networks with 50 epochs using AdamW optimizer, with a learning rate of $1e-4$ and weight decay of $1e-4$. For testing, we select the checkpoint with the best validation performance.

In HannesImitation, the model takes two consecutive images and joint positions as input. We train it to forecast eight future actions, and evaluate it by executing four future actions at test time. Unless otherwise specified, the remaining settings are identical to HannesImitation \cite{alessi2025hannesimitation}.

\subsection{Testing procedure}
During testing, we follow the same reach-to-grasp execution procedure as in data collection, with two modifications:
(1) the \emph{Stopping \& Grasping} phase is extended to 30 time steps; 
(2) finger motions are driven by the imitation-learning policy, while the wrist follows the reaching and lifting motions in collected dataset.

Although the demonstrations complete the grasp within 15 time steps, policies learned via imitation learning may require a longer execution horizon to finish finger closure and achieve a stable grasp. Using the original 15-step horizon would prematurely terminate some otherwise successful attempts and thus underestimate performance. Therefore, during testing we extend the \emph{Stopping \& Grasping} phase to 30 time steps, which we empirically found sufficient for grasp completion in simulation. This modification only increases the grasping completion window and keeps other components of the reach-and-lift procedure unchanged. In real-world use, users often wait until the grasp is completed before moving the prosthetic hand, making the extended horizon reasonable in practical interaction.
A dynamic/longer time window for the \emph{Stopping \& Grasping} phase may further improve grasp success; however, to keep the evaluation protocol simple and controlled, we do not investigate it in this work and leave it for future study.

\subsection{Evaluation} \label{sec:sim_eval}
We use the following criteria to assess grasp success and grasping behavior: (i) whether the policy can successfully grasp and stably lift the object, and (ii) whether it exhibits appropriate finger actions: keeping the hand open when it should not grasp and closing the fingers when it should grasp.

\textbf{Success rate (SR).} The ratio of successful trials to all testing trials. A trial is success, if $\left\| (p_{\mathrm{obj}}-p_{\mathrm{hand}})^{\mathrm{final}} - (p_{\mathrm{obj}}-p_{\mathrm{hand}})^{\mathrm{lift}} \right\|_2 < 0.03\,\mathrm{m}$, where $p_{\mathrm{hand}}$ denotes the position of the hand, and $p_{\mathrm{obj}}$ denotes the position of the object.
We empirically choose $\mathrm{lift}=85$ and $\mathrm{final}=116$, as these two time steps approximately correspond to the beginning and the end of the lifting phase, during which the relative position between the hand and the object remains nearly unchanged.

\textbf{Open during reaching rate (OR).} The ratio of trials that are classified as \textit{open during reach} among all test trials.
A trial is classified as \textit{open during reach} if no finger exceeds its corresponding threshold at any time step during the reaching phase, i.e.,
$
\sum_{t=1}^{T}\sum_{j=1}^{6}\mathbf{1}\bigl(|f_{t,j}|>\theta_t\bigr)=0.
$
Here, \(T\) denotes the final time step considered for evaluation and is empirically set to 40. \(f_{t,j}\) denotes the target value of finger \(j\in\{1,\dots,6\}\) at time step \(t\). \(\theta_t\) is a time-dependent threshold that increases linearly from 0.25 at \(t=1\) to 0.30 at \(t=T\). \(\mathbf{1}(\cdot)\) denotes the indicator function, which equals 1 if the condition inside is true, and 0 otherwise.


\textbf{Close before lifting rate (CR).} 
The ration of trials that are classified as \emph{close before lifting} to \emph{open during reach} trails.
A trial is classified as \emph{close before lifting} if $\sum_{j=1}^{6}\mathbf{1}\!\left(\left|f_{t,j} \right|> \theta_t\right) > 3$.
We empirically set $t=80$ and $\theta_{80}=0.4$.


\subsection{Generalization} \label{sec: sim_generalization}

In this section, we test generalization of imitation learning model \textbf{ACT} on unseen objects and/or environments.

\begin{figure*}[htbp]
  \centering  \includegraphics[page=1,width=\linewidth]{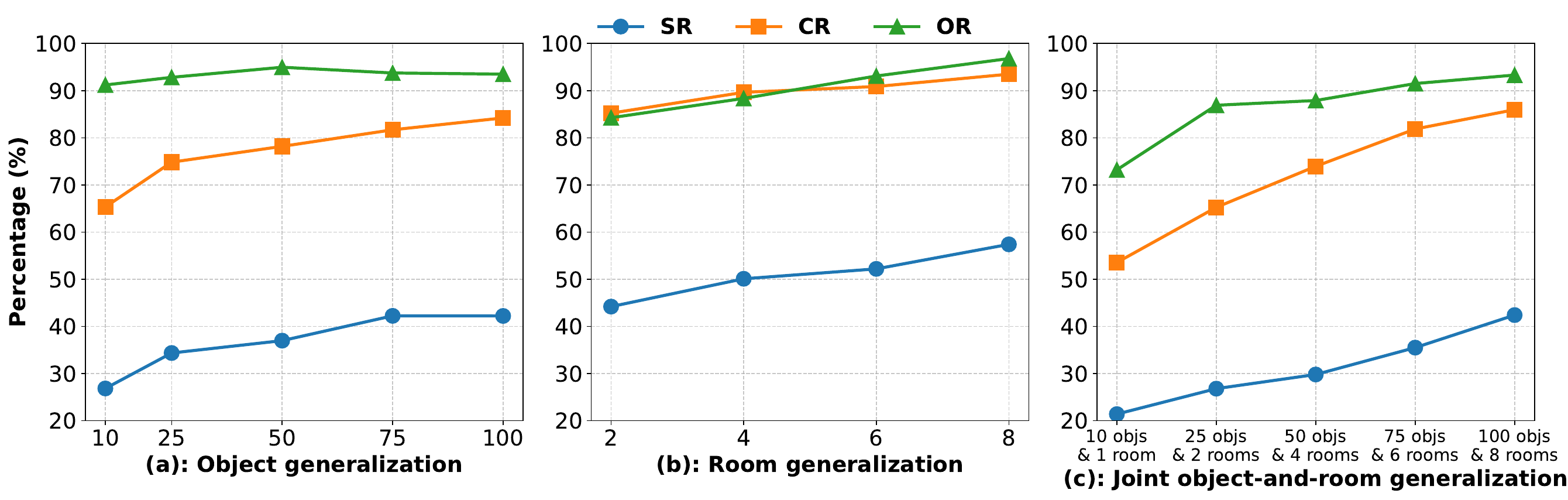}
  \caption{\textcolor{black}{Generalization performance for models trained on diverse number of objects, and rooms. 'SR' means success rate, 'CR' means close before lifting rate, and 'OR' means open during reaching rate.}}
  \label{fig:sim_generalization}
\end{figure*}

\textbf{Generalization settings.}
We evaluate generalization under three split settings.


\textit{(1) Object generalization.} We compare models trained on datasets with different numbers of objects, using 10, 25, 50, 75, or 100 objects, by evaluating all models on the same set of 100 test objects. For both training and testing, each object is associated with one trial in each room, resulting in 10 trials per object.


\textit{(2) Room generalization.} We compare models trained on datasets with different numbers of rooms, using 2, 4, 6, or 8 rooms, by evaluating all models on the same set of 2 test rooms. For both training and testing, each room is associated with 200 objects, resulting in 200 trials per room.


\textit{(3) Object \& room generalization.} We compare models trained on datasets with different numbers of objects and rooms, ranging from 10 objects in 1 room to 100 objects in 8 rooms, by evaluating all models on the same test set of 100 test objects in 2 test rooms.

\textbf{Object generalization.} We gradually expand seen (training) objects from 10 to 100, while the unseen (testing) objects are keep at 100.
Overall, as shown in Fig.~\ref{fig:sim_generalization}(a), increasing object diversity in training improves generalization: \emph{SR} rises from 26.84\% (train on 10 objects) to 42.22\% (train on 100 objects), and \emph{CR} increases from 65.33\% to 84.21\%. We observe diminishing returns beyond 75 training objects, where SR saturates (42.24\% $\rightarrow$ 42.22\%). In contrast, \emph{OR} remains consistently high (91.2--93.5\%) across all settings, suggesting that this metric is less sensitive to training object count compared to \emph{SR} and \emph{CR}.


\textbf{Room generalization.}
Fig.~\ref{fig:sim_generalization}(b) evaluates generalization to unseen rooms when training with different numbers of rooms.
Increasing the number of training environments consistently improves performance across all metrics:
\emph{SR} rises from 44.20\% (train on 2 rooms) to 57.40\% (train on 8 rooms), while \emph{CR} improves from 85.22\% to 93.49\%.
Similarly, \emph{OR} increases from 84.25\% to 96.80\%.
These results suggest that greater environmental diversity substantially enhances robustness to novel scene layouts and backgrounds.


\textbf{Joint object-and-room generalization.}
Fig.~\ref{fig:sim_generalization}(c) reports performance when simultaneously scaling both object and room diversity (from 10 objects/1 room to 100 objects/8 rooms) and evaluating on unseen objects and unseen rooms.
We observe a clear monotonic improvement with increasing diversity:
\emph{SR} improves from 21.40\% to 42.40\%, \emph{CR} from 53.55\% to 85.96\%, and \emph{OR} from 73.20\% to 93.30\%.
Notably, the absolute SR gains are larger than those observed when varying only the number of objects or only the number of rooms, indicating that object and environment diversity provide complementary benefits for generalization.


Across all three settings, increasing training diversity consistently improves generalization to unseen test conditions.
Expanding the number of training objects yields clear gains on unseen objects, with diminishing returns beyond $\sim$75 objects, while expanding the number of training rooms substantially boosts robustness to unseen rooms.
When jointly scaling both object and room diversity, performance improves monotonically and the gains are larger than varying either factor alone, indicating complementary benefits from object and room diversity for robust reach-to-grasp generalization.

\subsection{Imitation learning algorithms comparison}

\begin{table}[htbp]
\centering
\caption{Performance comparison under different algorithm and generalization settings. Bold indicates the best performance.}
\begin{tabular}{|c|c|c|c|c|}
\hline
\textbf{Algorithm} & \textbf{Generalization} & \textbf{SR} & \textbf{CR} & \textbf{OR} \\
\hline
\multirow{3}{*}{ACT~\cite{zhao2023learning}}     & Room           & \textbf{57.40\%} & 93.49\% & \textbf{96.80\%} \\ \cline{2-5}
                         & Object         & 42.22\% & 84.21\% & \textbf{93.50\%} \\ \cline{2-5}
                         & Room \& Object & 42.40\% & 85.96\% & \textbf{93.90\%}   \\ \hline
\multirow{3}{*}{VTM-VAE~\cite{shi2025towards}} & Room           & 55.75\% & \textbf{95.49\%} & 95.25\% \\ \cline{2-5}
                         & Object         & \textbf{44.90\%} & \textbf{93.13\%} & 91.70\% \\ \cline{2-5}
                         & Room \& Object & \textbf{44.20\% }& \textbf{94.50\%} & 92.70\% \\ \hline
\multirow{3}{*}{HannesImitation~\cite{alessi2025hannesimitation}}      & Room           & 48.95\% & 75.91\% & 93.60\% \\ \cline{2-5}
                         & Object         & 33.24\% & 66.87\% & 87.06\% \\ \cline{2-5}
                         & Room \& Object & 40.10\% & 77.28\% & 83.20\% \\ \hline
\end{tabular} \label{tab:algo_comparison}
\end{table}

Table~\ref{tab:algo_comparison} compares three imitation-learning baselines under three generalization settings: unseen rooms, unseen objects, and jointly unseen rooms \& objects.
Overall, \textbf{ACT} achieves the best performance for \emph{room} generalization, obtaining an SR of 57.40\% with strong CR (93.49\%) and OR (96.80\%).
For \emph{object} generalization, \textbf{VTM-VAE} performs best, improving SR from 42.22\% (ACT) to 44.90\% and yielding substantially higher CR (93.13\% vs.\ 84.21\%).
Under the hardest setting (\emph{room \& object}), \textbf{ACT} and \textbf{VTM-VAE} are comparable in SR (42.40\% vs.\ 44.20\%), while \textbf{VTM-VAE} maintains markedly higher CR (94.50\% vs.\ 85.96\%) and similar OR (92.70\% vs.\ 93.90\%).
In contrast, the prosthetic-specific diffusion baseline \textbf{HannesImitation} underperforms across all settings, especially on unseen objects (SR 33.24\%), indicating that robust generalization benefits from both action chunking and efficient visuomotor representation learning.

Overall, our simulation-collected demonstrations enable several representative SOTA imitation-learning policies to achieve strong performance in Isaac Lab, a physics-engine-based manipulation environment with rigid-body dynamics and contact interactions.
The non-trivial success rates and consistently high intermediate metrics across the evaluated generalization settings indicate that our dataset provides effective supervision for learning visuomotor prosthetic grasping behaviors in simulation.

\begin{table*}[htbp!]
\centering
\caption{Overall realistic-setting grasping performance. Note: 
SR denotes success rate; OR denotes Open During Hold rate; CR denotes Close during grasp rate (computed over OR trials). Object-level values are reported as mean $\pm$ standard deviation across participants. Stat-A aggregates across participants and objects, and Stat-B aggregates across participants, objects, and scenes.}
\begin{tabular}{|c|*{8}{c|}p{1.8cm}|} 
\hline
\textbf{Training Dataset} & \textbf{Scene} & \textbf{Criteria} &
\includegraphics[scale=0.2]{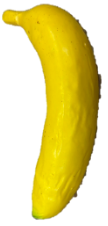} & 
\includegraphics[scale=0.2]{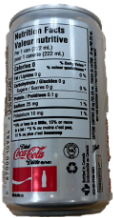} & 
\includegraphics[scale=0.06]{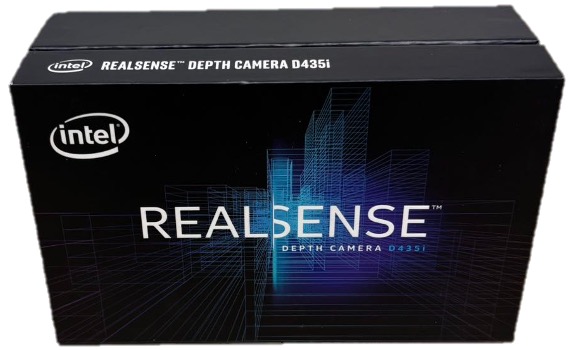}& 
\includegraphics[scale=0.2]{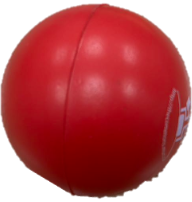}& 
\includegraphics[scale=0.03]{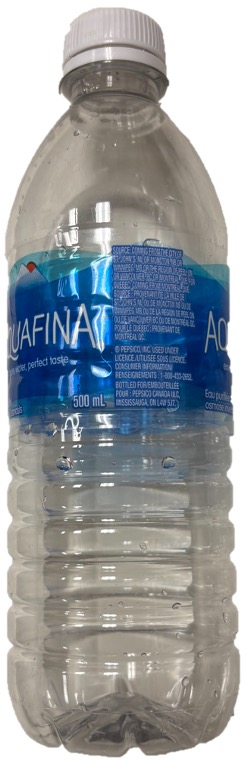} & \textbf{Stat-A} & \textbf{Stat-B}
\\  \cline{1-10}

\multirow{9}{*}{VTM-VAE~\cite{shi2025towards}} & \multirow{3}{*}{\includegraphics[scale=0.06]{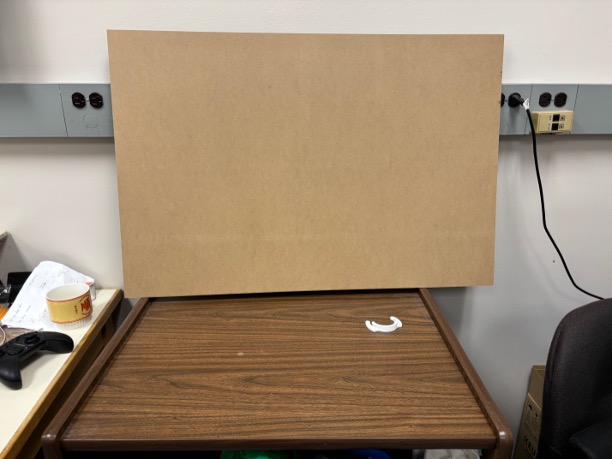}} & SR  &  $0.97_{\pm 0.07}$ & $1.00_{\pm 0.00}$ & $0.95_{\pm 0.09}$ & $1.00_{\pm 0.00}$ & $1.00_{\pm 0.00}$ &  $0.98_{\pm 0.06}$ & \multirow{9}{*}{%
\begin{tabular}{@{}l@{}}
    SR:   $0.66_{\pm 0.47}$\\
    OR: $0.67_{\pm 0.47}$\\
    CR: N/A
  \end{tabular}%
}\\ \cline{3-9}
& & OR & $1.00_{\pm 0.00}$ & $1.00_{\pm 0.00}$ & $1.00_{\pm 0.00}$ & $1.00_{\pm 0.00}$ & $1.00_{\pm 0.00}$ &  $1.00_{\pm 0.00}$ & \\ \cline{3-9}
& & CR &   $1.00_{\pm 0.00}$ & $1.00_{\pm 0.00}$ & $1.00_{\pm 0.00}$ & $1.00_{\pm 0.00}$ & $1.00_{\pm 0.00}$ &  $1.00_{\pm 0.00}$ & \\ \cline{2-9}

 & \multirow{3}{*}{\includegraphics[scale=0.022]{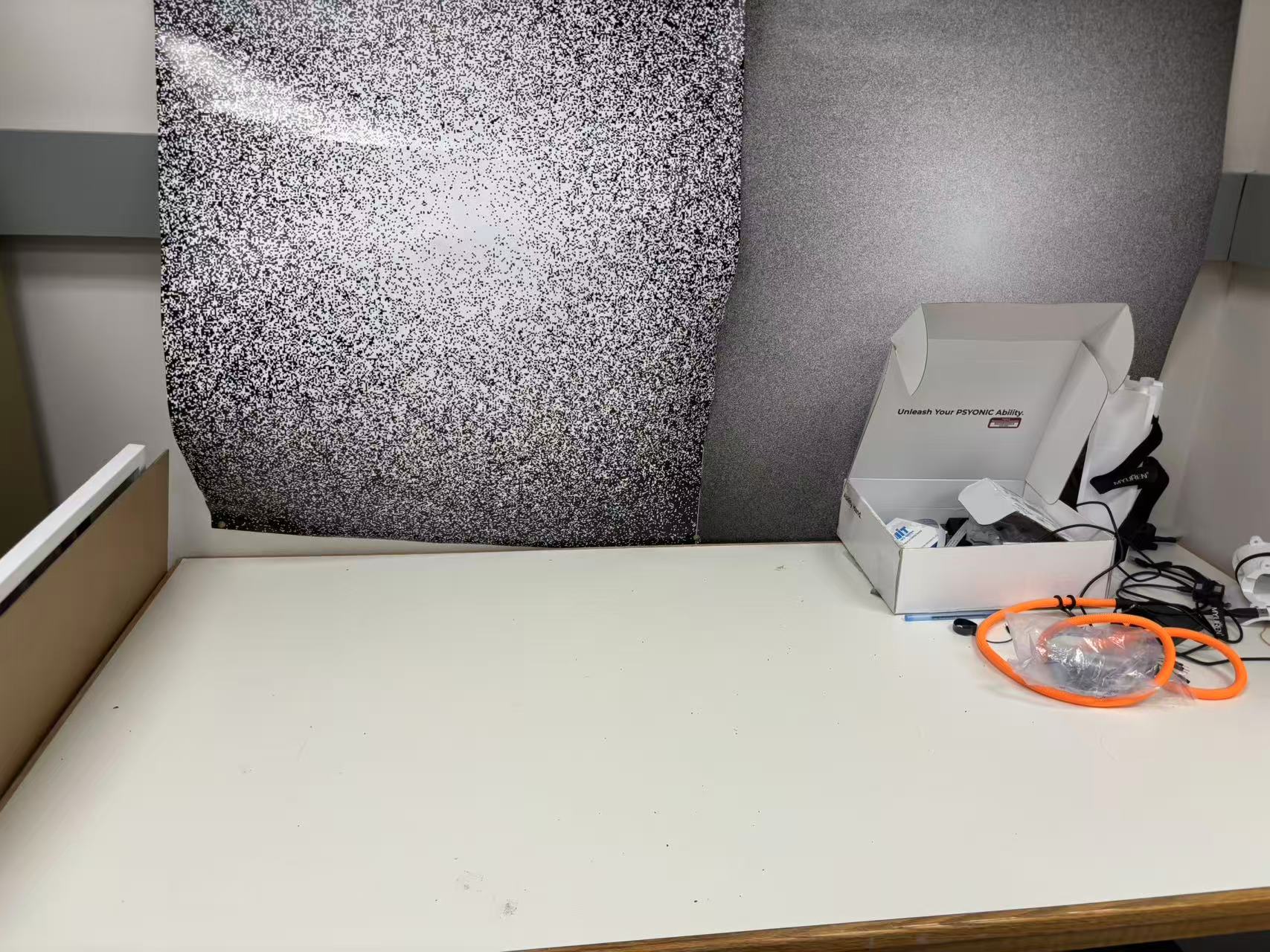}} & SR &   $1.00_{\pm 0.00}$ & $1.00_{\pm 0.00}$ & $0.97_{\pm 0.07}$ & $1.00_{\pm 0.00}$ & $1.00_{\pm 0.00}$ & $0.99_{\pm 0.04}$ &\\ \cline{3-9}
& & OR & $1.00_{\pm 0.00}$ & $1.00_{\pm 0.00}$ & $0.98_{\pm 0.06}$ & $1.00_{\pm 0.00}$ & $1.00_{\pm 0.00}$ & $0.99_{\pm 0.03}$&\\ \cline{3-9}
& & CR & $1.00_{\pm 0.00}$ & $1.00_{\pm 0.00}$ & $1.00_{\pm 0.00}$ & $1.00_{\pm 0.00}$ & $1.00_{\pm 0.00}$ & $1.00_{\pm 0.00}$ &\\ \cline{2-9}

 & \multirow{3}{*}{\includegraphics[scale=0.06]{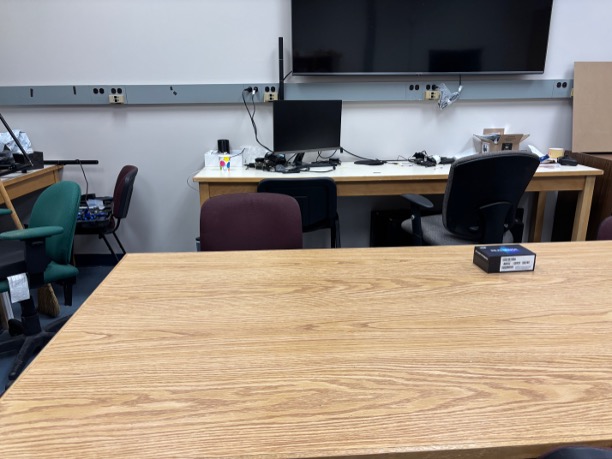}} & SR  & $0.00_{\pm 0.00}$ & $0.00_{\pm 0.00}$ & $0.00_{\pm 0.00}$ & $0.00_{\pm 0.00}$ & $0.00_{\pm 0.00}$ & $0.00_{\pm 0.00}$ & \\ \cline{3-9}
& & OR & $0.00_{\pm 0.00}$ & $0.00_{\pm 0.00}$ & $0.00_{\pm 0.00}$ & $0.00_{\pm 0.00}$ & $0.00_{\pm 0.00}$ & $0.00_{\pm 0.00}$ & \\ \cline{3-9}
& & CR & N/A & N/A &N/A  & N/A & N/A & N/A & \\ \cline{1-10}

\multirow{9}{*}{Ours} & \multirow{3}{*}{\includegraphics[scale=0.06]{figure/brown.jpg}} & SR & $0.97_{\pm 0.07}$ & $0.77_{\pm 0.27}$ & $0.82_{\pm 0.21}$ & $0.95_{\pm 0.12}$ & $0.98_{\pm 0.06}$ & $0.90_{\pm 0.19}$ &\multirow{9}{*}{%
\begin{tabular}{@{}l@{}}
    SR: $0.92_{\pm 0.15}$\\
    OR: $0.99_{\pm 0.01}$\\
    CR: $0.97_{\pm 0.09}$
  \end{tabular}%
}
\\ \cline{3-9}
& & OR &$1.00_{\pm 0.00}$ & $1.00_{\pm 0.00}$ & $1.00_{\pm 0.00}$ & $1.00_{\pm 0.00}$ & $1.00_{\pm 0.00}$ & $1.00_{\pm 0.00}$&\\ \cline{3-9}
& & CR & $0.98_{\pm 0.06}$ & $0.83_{\pm 0.23}$ & $0.92_{\pm 0.13}$ & $1.00_{\pm 0.00}$ & $1.00_{\pm 0.00}$ & $0.95_{\pm 0.14}$ & \\ \cline{2-9}

 & \multirow{3}{*}{\includegraphics[scale=0.022]{figure/white.jpg}} & SR   & $0.93_{\pm 0.09}$ & $0.88_{\pm 0.13}$ & $0.97_{\pm 0.07}$ & $0.95_{\pm 0.12}$ & $0.88_{\pm 0.24}$ & $0.92_{\pm 0.15}$ & \\ \cline{3-9}
& & OR & $1.00_{\pm 0.00}$ & $1.00_{\pm 0.00}$ & $1.00_{\pm 0.00}$ & $1.00_{\pm 0.00}$ & $1.00_{\pm 0.00}$ & $1.00_{\pm 0.00}$ &\\ \cline{3-9}
& & CR & $0.97_{\pm 0.07}$ & $0.97_{\pm 0.07}$ & $0.98_{\pm 0.06}$ & $1.00_{\pm 0.00}$ & $1.00_{\pm 0.00}$ & $0.98_{\pm 0.06}$  & \\ \cline{2-9}

& \multirow{3}{*}{\includegraphics[scale=0.06]{figure/yellow.jpg}} & SR  & $0.97_{\pm 0.07}$ & $0.97_{\pm 0.07}$ & $0.95_{\pm 0.12}$ & $0.92_{\pm 0.10}$ & $0.95_{\pm 0.09}$ & $0.95_{\pm 0.09}$&  \\ \cline{3-9}
& & OR & $0.98_{\pm 0.06}$ & $1.00_{\pm 0.00}$ & $1.00_{\pm 0.00}$ & $1.00_{\pm 0.00}$ & $1.00_{\pm 0.00}$ &$0.99_{\pm 0.03}$ & \\ \cline{3-9}
& & CR & $1.00_{\pm 0.00}$ & $0.98_{\pm 0.06}$ & $0.97_{\pm 0.07}$ & $0.98_{\pm 0.06}$ & $1.00_{\pm 0.00}$ & $0.99_{\pm 0.05}$& \\ \cline{1-10}

\end{tabular}
\label{tab:performance_real}
\end{table*}

\section{Realistic-setting Experiments}

We report realistic-setting experimental results from 12 participants included in the final analysis (13 enrolled, 1 withdrew), evaluated on 5 objects, 3 scenes, and 2 algorithms, with 5 recorded trials per object in each scene for each algorithm per participant (1800 trials in total).

\begin{figure}[htbp!]
  \centering  \includegraphics[page=1,width=\linewidth]{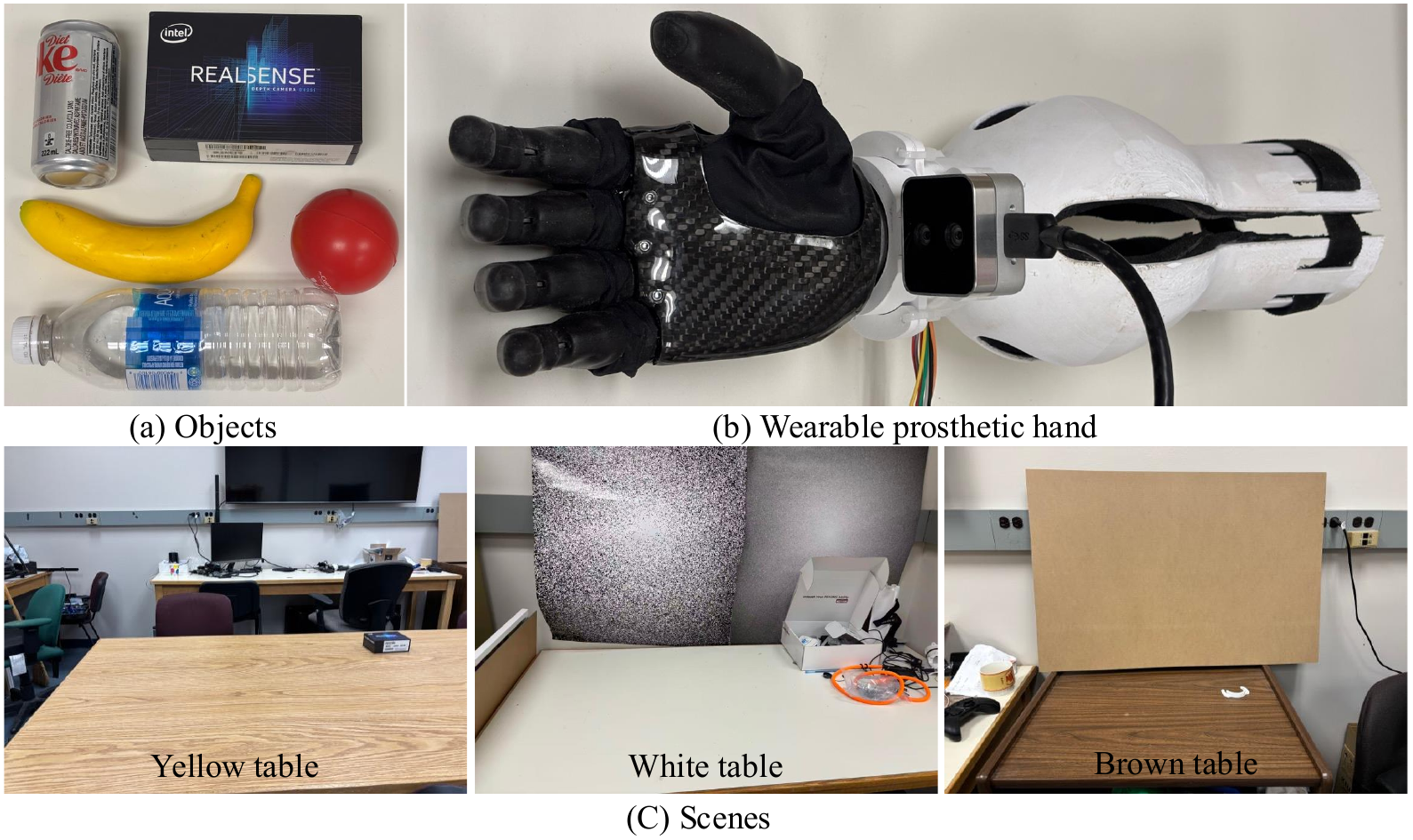}
  \caption{Objects, wearable prosthetic hand, and three scenes used in the realistic-setting experiments.}
  \label{fig:objectsHandReal}
\end{figure}

\subsection{Objects, prosthetic hand, and scenes}

\textbf{Objects.}
Fig.~\ref{fig:objectsHandReal}(a) shows the five objects used in the realistic-setting experiments (bottle, banana, ball, can, and box). These objects span diverse shapes, sizes, weights, and surface textures to cover a range of grasping conditions.

\textbf{Prosthetic hand.}
Fig.~\ref{fig:objectsHandReal}(b) shows the wearable prosthetic hand used in the experiments. The experimental setup follows \cite{shi2025towards}, including the same wearable hardware, camera and control interface.

\textbf{Scenes.} 
Fig.~\ref{fig:objectsHandReal}(c) shows the three experimental scenes, which vary in tabletop color and background wall texture. These scene variations are introduced to evaluate grasp robustness and to test whether the algorithm triggers unintended finger closure when no graspable object is present.

\subsection{Participants}

Thirteen healthy right-handed adults were enrolled in the study. One participant withdrew from the experiment and was excluded from subsequent analyses. Therefore, data from 12 participants (9 men and 3 women) were included in the final analysis.
Participants were 24--36 years old (\(M = 27.9\), \(SD = 3.9\)) and  157 to 183~cm tall (\(M = 171.6\), \(SD = 8.3\)).
We did not recruit amputee participants, as prior studies \cite{shi2025towards} have found that biosignals-free methods do not differ substantially between amputee and non-amputee conditions.

Because the number of trials was large, testing was conducted across two separate sessions. In each session, objects were randomly selected, and each session lasted approximately 1–1.5 hours. The interval between sessions ranged from 1--21~days (\(M = 6.3\), \(SD = 5.9\)).

Before data collection, participants watched an instructional video demonstrating the hold–grasp–lift sequence to familiarize them with the prosthetic control procedure. They then completed several practice trials (typically fewer than 10) before performing five consecutive recorded trials for each object. Practice trials were not included in the analysis.

The protocol was approved by the Memorial University Interdisciplinary Committee on Ethics in Human Research (20210316-SC), and written informed consent was obtained from all participants before participation.




\subsection{Inference Models for the Realistic Setting}
We evaluate two inference models under the realistic setting: a baseline trained on real-world data and our model trained entirely in simulation.

\textbf{VTM-VAE \cite{shi2025towards}.}
We train VTM-VAE using only the real-world grasping dataset from \cite{shi2025towards}, which was collected exclusively in white-table scenes (Fig.~\ref{fig:objectsHandReal}(c)).

\textbf{Ours.}
Without any real-world fine-tuning, we directly deploy the best-performing simulation-trained model from room generalization.

\subsection{Realistic-setting inference procedure}
In each trial, the computer provides three voice prompts in sequence: ``hold", ``grasp", and ``fail". The delay is 2 seconds between ``hold" and ``grasp", and 5 seconds between ``grasp" and ``fail".
\begin{enumerate} 
\item Hold: During the hold phase, the participant keeps the prosthetic hand approximately parallel to the tabletop. 
\item Grasp: The participant moves their right arm toward the target object. Once the prosthetic hand is properly positioned, the participant waits for the autonomous algorithm to execute the grasp. If the participant judges that the object has been successfully grasped, they lift the object; otherwise, they continue waiting until finger closure is complete or the fail prompt.
\item Fail: If the “fail” prompt is issued before the object is lifted, the trial is terminated and counted as a failure.
\end{enumerate}

\subsection{Evaluation criteria}

The criteria used in the simulation experiment (Sec.~\ref{sec:sim_eval}) were taken as a reference and adjusted to better represent realistic settings.

\textbf{Success rate (SR).} The ratio of successful grasp trials to total trials. A trial is considered successful if the participant lifts the object during the grasp phase.

\textbf{Open during hold rate (OR).} The ratio of trials in which the fingers remain open throughout the hold phase.

\textbf{Close during grasp rate (CR).} Among trials that satisfy Open During Hold, the ratio of trials in which the fingers close during the grasp phase.

\textbf{Definition of finger closure and opening.}
If three or more finger joints have angles greater than 20 degrees, the hand is classified as closed; otherwise, it is classified as open.

\subsection{Results}

Table \ref{tab:performance_real} summarizes the performance of the two methods in realistic settings across three scenes, five objects, and 12 participants, using the three criteria mentioned above. Object-wise performance are reported as mean $\pm$ standard deviation across participants. Stat-A summarizes performance by averaging across participants and objects, and Stat-B summarizes overall performance by averaging across participants, objects, and scenes.

\textbf{Overall performance.} The main result is that the proposed simulation-trained room-generalization model achieves substantially better overall robustness across scenes. Our method obtains a Stat-B SR of $0.92 _{\pm0.15}$, compared with $0.66 _{\pm 0.47}$ for VTM-VAE. A similar trend is observed for hold-phase stability: our method achieves OR = $0.99 _{\pm 0.01}$, whereas VTM-VAE achieves OR = $0.67 _{\pm 0.47}$. For CR (computed over OR trials), our method achieves $0.97 _{\pm 0.09}$, while VTM-VAE is N/A overall because it completely fails in one scene (yielding no valid OR trials for CR computation in that scene).

\textbf{Success rate.}
VTM-VAE performs near perfectly in the first two scenes (brown and white), with Stat-A SR values of $0.98 _{\pm 0.06}$ and $0.99 _{\pm 0.04}$, respectively. However, in the third scene (yellow), VTM-VAE collapses to $0.00 _{\pm 0.00}$ SR across all five objects. This failure pattern indicates strong sensitivity to scene appearance changes (background texture).

In contrast, our method maintains high SR in all three scenes, with Stat-A values of $0.90 _{\pm 0.19}$, $0.92 _{\pm 0.15}$, and $0.95 _{\pm 0.09}$. Although our method is slightly below VTM-VAE in the first two scenes, it remains robust in the third scene, where VTM-VAE fails completely. This cross-scene consistency is the primary source of the large improvement in overall Stat-B SR.

\textbf{Open during hold rate.}
For OR, VTM-VAE again performs well in the first two scenes (Stat-A: $1.00 _{\pm 0.00}$ and $0.99 _{\pm 0.03}$) but drops to $0.00 _{\pm 0.00}$ in the third scene. This indicates that the baseline cannot reliably maintain an open hand during the hold phase under the shifted scene condition.

Our method maintains near-ceiling OR in all scenes, with Stat-A values of $1.00 _{\pm 0.00}$, $1.00 _{\pm 0.00}$, and $0.99 _{\pm 0.03}$. These results show that the proposed method preserves stable hold-phase behavior under scene variation and avoids scene-induced false triggering.

\textbf{Close during grasp rate.}
CR is computed over trials satisfying OR and measures whether finger closure occurs during the grasp phase. VTM-VAE achieves CR = $1.00 _{\pm 0.00}$ in the first two scenes, but CR is N/A in the third scene because no trials satisfy OR there for calculation.

Our method achieves high and consistent CR across all scenes, with Stat-A values of $0.95 _{\pm 0.14}$, $0.98 _{\pm 0.06}$, and $0.99 _{\pm 0.05}$. This indicates that once a stable hold is established, the method reliably transitions to timely finger closure across different scene appearances.

\textbf{Scene-level interpretation.}
The dominant performance difference between the two methods is driven by scene variation rather than object identity. VTM-VAE performs strongly in two scenes but fails uniformly across all objects in the third scene, suggesting a severe visual domain shift. In contrast, our method remains effective in all three scenes, which is consistent with the use of procedurally varied simulation environments during training.

\textbf{Object-level trends.}
For our method, SR remains high across all five objects in all scenes, with moderate variation depending on the object-scene combination (e.g., lower SR for some objects in the first and second scenes). Importantly, these fluctuations remain localized and do not cause scene-wide failure. By contrast, the VTM-VAE degradation in the third scene affects all objects simultaneously, indicating poor scene-level generalization.

Overall, these results demonstrate that simulation-based room generalization model improves realistic-setting robustness to scene appearance changes, even if it yields slightly lower in-domain performance in easier scenes.

\subsection{Failure mode analysis}

\begin{table}[htbp]
  \centering
  \caption{Breakdown of failure modes for unsuccessful realistic-setting grasp trials.}
  \begin{tabular}{|l|c|}
    \hline
    \textbf{Failure mode} & \textbf{Count} \\ \hline
    No finger closure during grasp phase & 25 \\ \hline
    Insufficient grip force / slip & 32 \\ \hline
    Object pushed away during grasp & 8 \\ \hline
    Poor prosthetic hand pose & 3 \\ \hline
    Premature closure during hold phase & 1 \\ \hline
  \end{tabular} \label{tab:failure_grasp} 
\end{table}

To better understand the limitations of realistic-setting deployment, we categorize all unsuccessful trials into five failure modes (Table \ref{tab:failure_grasp}). The most common failures are insufficient grip force / slip (32 cases) and no finger closure during the grasp phase (25 cases), which together account for the majority of failures. These outcomes suggest that failures are dominated by either (i) missed or delayed grasp triggering (no closure), or (ii) insufficient grasp quality after closure (unstable grasp leading to slip or loss).

Less frequent failure modes include the object being pushed away during approach/contact (8 cases) and poor hand pose/misalignment (3 cases), indicating that approach trajectory and end-effector alignment can still lead to failures even when the grasp controller behaves as intended. Finally, premature closure during the hold phase is rare (1 case), consistent with the high OR values reported in Table \ref{tab:performance_real} and suggesting that unintended closure during hold is not a major issue in our setup.

\section{Conclusion and future work}

In this work, we presented a simulation-driven imitation learning framework for biosignals-free shared-autonomy prosthetic grasping.
By automatically generating diverse reach-to-grasp demonstrations in simulation, the proposed pipeline reduces the reliance on large-scale real-world human demonstrations and enables scalable policy training for sim-to-real transfer.
Experimental results in three real-world scenarios showed that the learned policy achieved grasping success rates of over 90\% and demonstrated stronger generalization than baseline methods. These findings indicate that simulation-driven imitation learning is a promising and practical direction for developing scalable biosignals-free shared-autonomy prosthetic grasping systems.
Future work will expand the dataset in terms of object and environment diversity, incorporate richer interaction dynamics such as multi-object clutter and articulated objects, improve sim-to-real transfer through better calibration and domain adaptation, and generalize the framework to a broader range of prosthetic hands and amputation types.



\bibliography{egbib}

\begin{thebibliography}{10}
\providecommand{\url}[1]{#1}
\csname url@samestyle\endcsname
\providecommand{\newblock}{\relax}
\providecommand{\bibinfo}[2]{#2}
\providecommand{\BIBentrySTDinterwordspacing}{\spaceskip=0pt\relax}
\providecommand{\BIBentryALTinterwordstretchfactor}{4}
\providecommand{\BIBentryALTinterwordspacing}{\spaceskip=\fontdimen2\font plus
\BIBentryALTinterwordstretchfactor\fontdimen3\font minus \fontdimen4\font\relax}
\providecommand{\BIBforeignlanguage}[2]{{%
\expandafter\ifx\csname l@#1\endcsname\relax
\typeout{** WARNING: IEEEtran.bst: No hyphenation pattern has been}%
\typeout{** loaded for the language `#1'. Using the pattern for}%
\typeout{** the default language instead.}%
\else
\language=\csname l@#1\endcsname
\fi
#2}}
\providecommand{\BIBdecl}{\relax}
\BIBdecl

\bibitem{dhawan2019proprioceptive}
A.~S. Dhawan, B.~Mukherjee, S.~Patwardhan, N.~Akhlaghi, G.~Diao, G.~Levay, R.~Holley, W.~M. Joiner, M.~Harris-Love, and S.~Sikdar, ``Proprioceptive sonomyographic control: A novel method for intuitive and proportional control of multiple degrees-of-freedom for individuals with upper extremity limb loss,'' \emph{Scientific reports}, vol.~9, no.~1, p. 9499, 2019.

\bibitem{farina2014extraction}
D.~Farina, N.~Jiang, H.~Rehbaum, A.~Holobar, B.~Graimann, H.~Dietl, and O.~C. Aszmann, ``The extraction of neural information from the surface emg for the control of upper-limb prostheses: emerging avenues and challenges,'' \emph{IEEE transactions on neural systems and rehabilitation engineering}, vol.~22, no.~4, pp. 797--809, 2014.

\bibitem{cipriani2011online}
C.~Cipriani, C.~Antfolk, M.~Controzzi, G.~Lundborg, B.~Ros{\'e}n, M.~C. Carrozza, and F.~Sebelius, ``Online myoelectric control of a dexterous hand prosthesis by transradial amputees,'' \emph{IEEE Transactions on Neural Systems and Rehabilitation Engineering}, vol.~19, no.~3, pp. 260--270, 2011.

\bibitem{roche2014prosthetic}
A.~D. Roche, H.~Rehbaum, D.~Farina, and O.~C. Aszmann, ``Prosthetic myoelectric control strategies: a clinical perspective,'' \emph{Current Surgery Reports}, vol.~2, pp. 1--11, 2014.

\bibitem{nasr2021musclenet}
A.~Nasr, S.~Bell, J.~He, R.~L. Whittaker, N.~Jiang, C.~R. Dickerson, and J.~McPhee, ``Musclenet: mapping electromyography to kinematic and dynamic biomechanical variables by machine learning,'' \emph{Journal of Neural Engineering}, vol.~18, no.~4, p. 0460d3, 2021.

\bibitem{boostani2003evaluation}
R.~Boostani and M.~H. Moradi, ``Evaluation of the forearm emg signal features for the control of a prosthetic hand,'' \emph{Physiological Measurement}, vol.~24, no.~2, p. 309, 2003.

\bibitem{dalley2011method}
S.~A. Dalley, H.~A. Varol, and M.~Goldfarb, ``A method for the control of multigrasp myoelectric prosthetic hands,'' \emph{IEEE Transactions on Neural Systems and Rehabilitation Engineering}, vol.~20, no.~1, pp. 58--67, 2011.

\bibitem{farrell2007optimal}
T.~R. Farrell and R.~F. Weir, ``The optimal controller delay for myoelectric prostheses,'' \emph{IEEE Transactions on neural systems and rehabilitation engineering}, vol.~15, no.~1, pp. 111--118, 2007.

\bibitem{wang2021effect}
J.~Wang, M.~Pang, P.~Yu, B.~Tang, K.~Xiang, and Z.~Ju, ``Effect of muscle fatigue on surface electromyography-based hand grasp force estimation,'' \emph{Applied Bionics and Biomechanics}, vol. 2021, no.~1, p. 8817480, 2021.

\bibitem{fang2022simultaneous}
B.~Fang, C.~Wang, F.~Sun, Z.~Chen, J.~Shan, H.~Liu, W.~Ding, and W.~Liang, ``Simultaneous semg recognition of gestures and force levels for interaction with prosthetic hand,'' \emph{IEEE Transactions on Neural Systems and Rehabilitation Engineering}, vol.~30, pp. 2426--2436, 2022.

\bibitem{dovsen2010cognitive}
S.~Do{\v{s}}en, C.~Cipriani, M.~Kosti{\'c}, M.~Controzzi, M.~C. Carrozza, and D.~B. Popovi{\'c}, ``Cognitive vision system for control of dexterous prosthetic hands: experimental evaluation,'' \emph{Journal of Neuroengineering and Rehabilitation}, vol.~7, pp. 1--14, 2010.

\bibitem{he2020vision}
Y.~He, R.~Kubozono, O.~Fukuda, N.~Yamaguchi, and H.~Okumura, ``Vision-based assistance for myoelectric hand control,'' \emph{IEEE Access}, vol.~8, pp. 201\,956--201\,965, 2020.

\bibitem{shi2025towards}
K.~Shi, W.~Lu, H.~Zhao, V.~Prado~da Fonseca, T.~Zou, and X.~Jiang, ``Toward biosignals-free autonomous prosthetic hand control via imitation learning,'' \emph{IEEE Transactions on Neural Systems and Rehabilitation Engineering}, vol.~33, pp. 3544--3554, 2025.

\bibitem{alessi2025hannesimitation}
C.~Alessi, F.~Vasile, F.~Ceola, G.~Pasquale, N.~Boccardo, and L.~Natale, ``Hannesimitation: Grasping with the hannes prosthetic hand via imitation learning,'' in \emph{2025 IEEE/RSJ International Conference on Intelligent Robots and Systems (IROS)}.\hskip 1em plus 0.5em minus 0.4em\relax IEEE, 2025, pp. 10\,085--10\,092.

\bibitem{gupta2018robot}
A.~Gupta, A.~Murali, D.~P. Gandhi, and L.~Pinto, ``Robot learning in homes: Improving generalization and reducing dataset bias,'' \emph{Advances in neural information processing systems}, vol.~31, 2018.

\bibitem{tobin2017domain}
J.~Tobin, R.~Fong, A.~Ray, J.~Schneider, W.~Zaremba, and P.~Abbeel, ``Domain randomization for transferring deep neural networks from simulation to the real world,'' in \emph{2017 IEEE/RSJ international conference on intelligent robots and systems (IROS)}.\hskip 1em plus 0.5em minus 0.4em\relax IEEE, 2017, pp. 23--30.

\bibitem{hsu2022vision}
K.~Hsu, M.~J. Kim, R.~Rafailov, J.~Wu, and C.~Finn, ``Vision-based manipulators need to also see from their hands.'' in \emph{ICLR}, 2022.

\bibitem{kim24openvla}
M.~Kim, K.~Pertsch, S.~Karamcheti, T.~Xiao, A.~Balakrishna, S.~Nair, R.~Rafailov, E.~Foster, G.~Lam, P.~Sanketi, Q.~Vuong, T.~Kollar, B.~Burchfiel, R.~Tedrake, D.~Sadigh, S.~Levine, P.~Liang, and C.~Finn, ``Openvla: An open-source vision-language-action model,'' \emph{arXiv preprint arXiv:2406.09246}, 2024.

\bibitem{xie2021virtual}
J.~Xie and X.~Hu, ``Virtual reality for evaluating prosthetic hand control strategies: A preliminary report,'' in \emph{2021 43rd Annual International Conference of the IEEE Engineering in Medicine \& Biology Society (EMBC)}.\hskip 1em plus 0.5em minus 0.4em\relax IEEE, 2021, pp. 6263--6266.

\bibitem{geethanjali2014low}
P.~Geethanjali and K.~Ray, ``A low-cost real-time research platform for emg pattern recognition-based prosthetic hand,'' \emph{IEEE/ASME Transactions on Mechatronics}, vol.~20, no.~4, pp. 1948--1955, 2014.

\bibitem{patel2018classification}
G.~K. Patel, C.~Castellini, J.~M. Hahne, D.~Farina, and S.~Dosen, ``A classification method for myoelectric control of hand prostheses inspired by muscle coordination,'' \emph{IEEE Transactions on Neural Systems and Rehabilitation Engineering}, vol.~26, no.~9, pp. 1745--1755, 2018.

\bibitem{hahne2015concurrent}
J.~M. Hahne, S.~D{\"a}hne, H.-J. Hwang, K.-R. M{\"u}ller, and L.~C. Parra, ``Concurrent adaptation of human and machine improves simultaneous and proportional myoelectric control,'' \emph{IEEE Transactions on Neural Systems and Rehabilitation Engineering}, vol.~23, no.~4, pp. 618--627, 2015.

\bibitem{jiang2012myoelectric}
N.~Jiang, S.~Dosen, K.-R. Muller, and D.~Farina, ``Myoelectric control of artificial limbs—is there a need to change focus?[in the spotlight],'' \emph{IEEE Signal Processing Magazine}, vol.~29, no.~5, pp. 152--150, 2012.

\bibitem{jiang2013intuitive}
N.~Jiang, H.~Rehbaum, I.~Vujaklija, B.~Graimann, and D.~Farina, ``Intuitive, online, simultaneous, and proportional myoelectric control over two degrees-of-freedom in upper limb amputees,'' \emph{IEEE transactions on neural systems and rehabilitation engineering}, vol.~22, no.~3, pp. 501--510, 2013.

\bibitem{xu2025powered}
Y.~Xu, X.~Wang, J.~Li, X.~Zhang, F.~Li, Q.~Gao, C.~Fu, and Y.~Leng, ``A powered prosthetic hand with vision system for enhancing the anthropopathic grasp,'' \emph{IEEE Transactions on Neural Systems and Rehabilitation Engineering}, 2025.

\bibitem{montagnani2015exploiting}
F.~Montagnani, M.~Controzzi, and C.~Cipriani, ``Exploiting arm posture synergies in activities of daily living to control the wrist rotation in upper limb prostheses: A feasibility study,'' in \emph{2015 37th Annual International Conference of the IEEE Engineering in Medicine and Biology Society}.\hskip 1em plus 0.5em minus 0.4em\relax IEEE, 2015, pp. 2462--2465.

\bibitem{kuhn2024synergy}
J.~K{\"u}hn, T.~Hu, A.~T{\"o}dtheide, E.~Pozo~Fortuni{\'c}, E.~Jensen, and S.~Haddadin, ``The synergy complement control approach for seamless limb-driven prostheses,'' \emph{Nature Machine Intelligence}, vol.~6, no.~4, pp. 481--492, 2024.

\bibitem{merad2020assessment}
M.~Merad, E.~De~Montalivet, M.~Legrand, E.~Mastinu, M.~Ortiz-Catalan, A.~Touillet, N.~Martinet, J.~Paysant, A.~Roby-Brami, and N.~Jarrasse, ``Assessment of an automatic prosthetic elbow control strategy using residual limb motion for transhumeral amputated individuals with socket or osseointegrated prostheses,'' \emph{IEEE Transactions on Medical Robotics and Bionics}, vol.~2, no.~1, pp. 38--49, 2020.

\bibitem{bennett2017imu}
D.~A. Bennett and M.~Goldfarb, ``Imu-based wrist rotation control of a transradial myoelectric prosthesis,'' \emph{IEEE Transactions on Neural Systems and Rehabilitation Engineering}, vol.~26, no.~2, pp. 419--427, 2017.

\bibitem{pilarski2012dynamic}
P.~M. Pilarski, M.~R. Dawson, T.~Degris, J.~P. Carey, and R.~S. Sutton, ``Dynamic switching and real-time machine learning for improved human control of assistive biomedical robots,'' in \emph{2012 4th IEEE RAS \& EMBS International Conference on Biomedical Robotics and Biomechatronics}.\hskip 1em plus 0.5em minus 0.4em\relax IEEE, 2012, pp. 296--302.

\bibitem{mastinu2024explorations}
E.~Mastinu, A.~Coletti, J.~van~den Berg, and C.~Cipriani, ``Explorations of autonomous prosthetic grasping via proximity vision and deep learning,'' \emph{IEEE Transactions on Medical Robotics and Bionics}, 2024.

\bibitem{heo2023proximity}
S.-H. Heo and H.-S. Park, ``Proximity perception-based grasping intelligence: toward the seamless control of a dexterous prosthetic hand,'' \emph{IEEE/ASME Transactions on Mechatronics}, vol.~29, no.~3, pp. 2079--2090, 2023.

\bibitem{fang2020graspnet}
H.-S. Fang, C.~Wang, M.~Gou, and C.~Lu, ``Graspnet-1billion: A large-scale benchmark for general object grasping,'' in \emph{Proceedings of the IEEE/CVF conference on computer vision and pattern recognition}, 2020, pp. 11\,444--11\,453.

\bibitem{chen2025bodex}
J.~Chen, Y.~Ke, and H.~Wang, ``Bodex: Scalable and efficient robotic dexterous grasp synthesis using bilevel optimization,'' in \emph{2025 IEEE International Conference on Robotics and Automation (ICRA)}.\hskip 1em plus 0.5em minus 0.4em\relax IEEE, 2025, pp. 01--08.

\bibitem{wang2022dexgraspnet}
R.~Wang, J.~Zhang, J.~Chen, Y.~Xu, P.~Li, T.~Liu, and H.~Wang, ``Dexgraspnet: A large-scale robotic dexterous grasp dataset for general objects based on simulation,'' \emph{arXiv preprint arXiv:2210.02697}, 2022.

\bibitem{liu2024structured}
H.~Liu, H.~Li, C.~Jiang, S.~Xue, Y.~Zhao, X.~Huang, and Z.~Jiang, ``Structured local feature-conditioned 6-dof variational grasp detection network in cluttered scenes,'' \emph{IEEE/ASME Transactions on Mechatronics}, 2024.

\bibitem{yu2025trustworthy}
H.~Yu, X.~Zhang, Z.~Zhao, and C.~He, ``Trustworthy robotic grasping: A credibility alignment framework via self-regulation encoding,'' \emph{IEEE/ASME Transactions on Mechatronics}, 2025.

\bibitem{xu2025anthropomorphic}
W.~Xu, Z.~Geng, X.~Shi, W.~Guo, and X.~Sheng, ``Anthropomorphic grasp motion planning for humanoid robots via learned riemannian metric and dextrous grasp evaluator,'' \emph{IEEE/ASME Transactions on Mechatronics}, 2025.

\bibitem{wang2025toward}
Q.~Wang, K.~Bai, L.~Zhang, Q.~Li, A.~Knoll, J.~Zhang, Y.~Ying, and M.~Zhou, ``Toward collision-aware robotic fragile fruit grasping: A sim-to-real framework for perception, reasoning, and execution,'' \emph{IEEE/ASME Transactions on Mechatronics}, 2025.

\bibitem{zhang2024dexgraspnet}
J.~Zhang, H.~Liu, D.~Li, X.~Yu, H.~Geng, Y.~Ding, J.~Chen, and H.~Wang, ``Dexgraspnet 2.0: Learning generative dexterous grasping in large-scale synthetic cluttered scenes,'' in \emph{8th Annual Conference on Robot Learning}, 2024.

\bibitem{tobin2018domain}
J.~Tobin, L.~Biewald, R.~Duan, M.~Andrychowicz, A.~Handa, V.~Kumar, B.~McGrew, A.~Ray, J.~Schneider, P.~Welinder \emph{et~al.}, ``Domain randomization and generative models for robotic grasping,'' in \emph{2018 IEEE/RSJ International Conference on Intelligent Robots and Systems (IROS)}.\hskip 1em plus 0.5em minus 0.4em\relax IEEE, 2018, pp. 3482--3489.

\bibitem{huber2024domain}
J.~Huber, F.~H{\'e}l{\'e}non, H.~Watrelot, F.~B. Amar, and S.~Doncieux, ``Domain randomization for sim2real transfer of automatically generated grasping datasets,'' in \emph{2024 IEEE international conference on robotics and automation (ICRA)}.\hskip 1em plus 0.5em minus 0.4em\relax IEEE, 2024, pp. 4112--4118.

\bibitem{zhao2023learning}
T.~Z. Zhao, V.~Kumar, S.~Levine, and C.~Finn, ``Learning fine-grained bimanual manipulation with low-cost hardware,'' \emph{arXiv preprint arXiv:2304.13705}, 2023.

\bibitem{mandlekar2023human}
A.~Mandlekar, C.~R. Garrett, D.~Xu, and D.~Fox, ``Human-in-the-loop task and motion planning for imitation learning,'' in \emph{Conference on Robot Learning}.\hskip 1em plus 0.5em minus 0.4em\relax PMLR, 2023, pp. 3030--3060.

\bibitem{choudhury2018data}
S.~Choudhury, M.~Bhardwaj, S.~Arora, A.~Kapoor, G.~Ranade, S.~Scherer, and D.~Dey, ``Data-driven planning via imitation learning,'' \emph{The International Journal of Robotics Research}, vol.~37, no. 13-14, pp. 1632--1672, 2018.

\bibitem{zare2024survey}
M.~Zare, P.~M. Kebria, A.~Khosravi, and S.~Nahavandi, ``A survey of imitation learning: Algorithms, recent developments, and challenges,'' \emph{IEEE Transactions on Cybernetics}, vol.~54, no.~12, pp. 7173--7186, 2024.

\bibitem{chi2023diffusion}
C.~Chi, Z.~Xu, S.~Feng, E.~Cousineau, Y.~Du, B.~Burchfiel, R.~Tedrake, and S.~Song, ``Diffusion policy: Visuomotor policy learning via action diffusion,'' \emph{The International Journal of Robotics Research}, p. 02783649241273668, 2023.

\bibitem{torabi2018behavioral}
F.~Torabi, G.~Warnell, and P.~Stone, ``Behavioral cloning from observation,'' \emph{arXiv preprint arXiv:1805.01954}, 2018.

\bibitem{ross2011reduction}
S.~Ross, G.~Gordon, and D.~Bagnell, ``A reduction of imitation learning and structured prediction to no-regret online learning,'' in \emph{Proceedings of the fourteenth international conference on artificial intelligence and statistics}.\hskip 1em plus 0.5em minus 0.4em\relax JMLR Workshop and Conference Proceedings, 2011, pp. 627--635.

\bibitem{kelly2019hg}
M.~Kelly, C.~Sidrane, K.~Driggs-Campbell, and M.~J. Kochenderfer, ``Hg-dagger: Interactive imitation learning with human experts,'' in \emph{2019 International Conference on Robotics and Automation (ICRA)}.\hskip 1em plus 0.5em minus 0.4em\relax IEEE, 2019, pp. 8077--8083.

\bibitem{levine2016end}
S.~Levine, C.~Finn, T.~Darrell, and P.~Abbeel, ``End-to-end training of deep visuomotor policies,'' \emph{Journal of Machine Learning Research}, vol.~17, no.~39, pp. 1--40, 2016.

\bibitem{james2019sim}
S.~James, P.~Wohlhart, M.~Kalakrishnan, D.~Kalashnikov, A.~Irpan, J.~Ibarz, S.~Levine, R.~Hadsell, and K.~Bousmalis, ``Sim-to-real via sim-to-sim: Data-efficient robotic grasping via randomized-to-canonical adaptation networks,'' in \emph{Proceedings of the IEEE/CVF conference on computer vision and pattern recognition}, 2019, pp. 12\,627--12\,637.

\bibitem{patel2025real}
S.~Patel, X.~Yin, W.~Huang, S.~Garg, H.~Nayyeri, L.~Fei-Fei, S.~Lazebnik, and Y.~Li, ``A real-to-sim-to-real approach to robotic manipulation with vlm-generated iterative keypoint rewards,'' in \emph{2025 IEEE International Conference on Robotics and Automation (ICRA)}.\hskip 1em plus 0.5em minus 0.4em\relax IEEE, 2025, pp. 8258--8266.

\bibitem{ju2022transferring}
H.~Ju, R.~Juan, R.~Gomez, K.~Nakamura, and G.~Li, ``Transferring policy of deep reinforcement learning from simulation to reality for robotics,'' \emph{Nature Machine Intelligence}, vol.~4, no.~12, pp. 1077--1087, 2022.

\bibitem{ankile2025residual}
L.~Ankile, Z.~Jiang, R.~Duan, G.~Shi, P.~Abbeel, and A.~Nagabandi, ``Residual off-policy rl for finetuning behavior cloning policies,'' \emph{arXiv preprint arXiv:2509.19301}, 2025.

\bibitem{downs2022google}
L.~Downs, A.~Francis, N.~Koenig, B.~Kinman, R.~Hickman, K.~Reymann, T.~B. McHugh, and V.~Vanhoucke, ``Google scanned objects: A high-quality dataset of 3d scanned household items,'' in \emph{2022 International Conference on Robotics and Automation (ICRA)}.\hskip 1em plus 0.5em minus 0.4em\relax IEEE, 2022, pp. 2553--2560.

\bibitem{casas2024multigrippergrasp}
L.~F. Casas, N.~Khargonkar, B.~Prabhakaran, and Y.~Xiang, ``Multigrippergrasp: A dataset for robotic grasping from parallel jaw grippers to dexterous hands,'' in \emph{2024 IEEE/RSJ International Conference on Intelligent Robots and Systems (IROS)}.\hskip 1em plus 0.5em minus 0.4em\relax IEEE, 2024, pp. 2978--2984.

\bibitem{psyonic_touch_sensing_bionic_hand_2024}
{PSYONIC}, ``The world's first touch-sensing bionic hand,'' [Online]. Available: \url{https://www.psyonic.io/}, 2024, accessed: 2026-01-19.

\bibitem{raistrick2024infinigen}
A.~Raistrick, L.~Mei, K.~Kayan, D.~Yan, Y.~Zuo, B.~Han, H.~Wen, M.~Parakh, S.~Alexandropoulos, L.~Lipson \emph{et~al.}, ``Infinigen indoors: Photorealistic indoor scenes using procedural generation,'' in \emph{Proceedings of the IEEE/CVF Conference on Computer Vision and Pattern Recognition}, 2024, pp. 21\,783--21\,794.

\bibitem{chen2024bodex}
J.~Chen, Y.~Ke, and H.~Wang, ``Bodex: Scalable and efficient robotic dexterous grasp synthesis using bilevel optimization,'' \emph{arXiv preprint arXiv:2412.16490}, 2024.

\bibitem{stracquadanio2025bring}
G.~Stracquadanio, F.~Vasile, E.~Maiettini, N.~Boccardo, and L.~Natale, ``Bring your own grasp generator: Leveraging robot grasp generation for prosthetic grasping,'' \emph{arXiv preprint arXiv:2503.00466}, 2025.

\bibitem{sundermeyer2021contact}
M.~Sundermeyer, A.~Mousavian, R.~Triebel, and D.~Fox, ``Contact-graspnet: Efficient 6-dof grasp generation in cluttered scenes,'' in \emph{2021 IEEE International Conference on Robotics and Automation (ICRA)}.\hskip 1em plus 0.5em minus 0.4em\relax IEEE, 2021, pp. 13\,438--13\,444.

\bibitem{wu2023learning}
T.~Wu, M.~Wu, J.~Zhang, Y.~Gan, and H.~Dong, ``Learning score-based grasping primitive for human-assisting dexterous grasping,'' \emph{Advances in Neural Information Processing Systems}, vol.~36, pp. 22\,132--22\,150, 2023.

\bibitem{chao2021dexycb}
Y.-W. Chao, W.~Yang, Y.~Xiang, P.~Molchanov, A.~Handa, J.~Tremblay, Y.~S. Narang, K.~Van~Wyk, U.~Iqbal, S.~Birchfield \emph{et~al.}, ``Dexycb: A benchmark for capturing hand grasping of objects,'' in \emph{Proceedings of the IEEE/CVF conference on computer vision and pattern recognition}, 2021, pp. 9044--9053.

\end{thebibliography}
\bibliographystyle{IEEEtran}

\end{document}